# *Emotions are Universal:* Learning Sentiment Based Representations of Resource-Poor Languages using Siamese Networks

Nurendra Choudhary⋆, Rajat Singh⋆, Ishita Bindlish and Manish Shrivastava

Language Technologies Research Centre (LTRC)
Kohli Center on Intelligent Systems (KCIS)
International Institute of Information Technology, Hyderabad, India
{nurendra.choudhary,rajat.singh}@research.iiit.ac.in
ishita.bindlish@students.iiit.ac.in
m.shrivastava@iiit.ac.in

**Abstract** Machine learning approaches in sentiment analysis principally rely on the abundance of resources. To limit this dependence, we propose a novel method called *Siamese Network Architecture for Sentiment Analysis (SNASA)* to learn representations of resource-poor languages by jointly training them with resource-rich languages using a siamese network.
SNASA model consists of twin Bi-directional Long Short-Term Memory Recurrent Neural Networks (Bi-LSTM RNN) with shared parameters joined by a contrastive loss function, based on a similarity metric. The model learns the sentence representations of resource-poor and resource-rich language in a common sentiment space by using a similarity metric based on their individual sentiments. The model, hence, projects sentences with similar sentiment closer to each other and the sentences with different sentiment farther from each other. Experiments on large-scale datasets of resource-rich languages - English and Spanish and resource-poor languages - Hindi and Telugu reveal that SNASA outperforms the state-of-the-art sentiment analysis approaches based on distributional semantics, semantic rules, lexicon lists and deep neural network representations without shared parameters.

## 1 Introduction

With proliferation of the Internet into multilingual communities, the linguistic diversity of the real world is being reflected in the virtual world too. Opinionated data like reviews and recommendations are a crucial source of critical analysis for businesses looking for customer experience, expansion into a new segment or their general perception in the market. The data also significantly impacts political policies and campaigns as they represent the public perspective.

---

⋆ These authors have contributed equally to this work.

Sentiment analysis or polarity detection is a widely studied field in natural language processing with several approaches ranging from rule-based systems to deep learning architectures. Deep learning approaches proved exceptionally effective in solving the task. However, a primary component necessary for the effectiveness of these deep learning approaches is the abundance of data. Hence, major deep learning architectures do not yield satisfactory results in languages with scarce resources. Hence, to overcome the problem, we leverage the abundant resources available in other languages and map both the languages to a common sentiment space.

In this paper, we propose a unified architecture called *Siamese Network Architecture for Sentiment Analysis (SNASA)*. The model consists of twin bi-directional LSTM networks with shared parameters, joined together by a contrastive loss function. The energy function suits the discriminative training for energy based models [14].

SNASA model starts with a simple primary representation based on character trigrams. The model then learns the sentence representation by utilizing the similarity based contrastive energy function. The contrastive function maps the sentences into the sentiment space, such that the distance between sentences with same sentiment is minimized and distance between sentences with different sentiment is maximized. For example, "I am very happy." and "यह बहुत अच्छी किताब है"(This is a very good book) are closer to each other, whereas, "This is the worst day" and "बगीचा सुन्दर है"(The garden is beautiful) are farther from each other in the sentiment space.

SNASA is a siamese network with shared parameters. We utilize the shared parameters to learn the sentiment based representation for languages with poor resources by jointly training them with resource-rich languages. The model, thus, establishes a correlation between the resource-rich and resource-poor language and maps them to the same sentiment space. This correlation is further utilized to predict the sentiment of the resource-poor languages using the immense data available in resource-rich languages.

The rest of the paper is organized as follows. Section 2 presents the previous approaches to conquer the problem. Section 3 describes the evaluation dataset and section 4 describes the architecture of SNASA. Section 5 explains the training and testing phase of SNASA. Section 6 details the baselines. Section 7 presents the experimental set-up and results. Finally, section 8 concludes the paper.

## 2   Related Work

Sentiment analysis is a widely studied task with various approaches proposed in the recent period. In this section, we survey the previous methodologies for the task.

Distributional semantics [15] approach captures the sentence's overall semantic value but does not maintain information of the words' order. [18] propose classifier models based on support vector machines that assigns sentiment po-larity to words or phrases using classifiers. Polarity of its constituents totals the sentences' polarity. Lexicon based approaches [23] utilize a manually constructed lexicon with sentiments of major words given. This information assigns the polarity. The limitation of these approaches is the information loss of the words' sequence which leads to the wrong classification. e.g; In *"I am not happy"*, *"not"* carries a negative sentiment and *"happy"* carries a positive sentiment. The combination gives a neutral sentiment, whereas the sentence is truly negative. Bag of n-grams limit this effect but do not eliminate it completely.

Matrix Vector Recursive Neural Network (MV-RNN) [22] provides a solution to the problem of capturing the words' relation in a sentence. The model assigns a vector and a matrix to each node of the sentence's syntactically parsed tree. The vector and matrix represent the word's semantic value and its relation with the other words respectively. This approach presents an effective model for capturing the content and relations of the sentence. However, the approach requires a large amount of data to train and hence will fail in case of languages with fewer resources.

Adaboost based Convolution Neural Networks (Ada-CNN) [10] uses CNN classifiers with different filter sizes. Adaboost arrives at a weighted combination of the classifiers. The differing filter sizes analyze the contribution of different n-grams to the overall sentiment.

Another line of research [11,2] utilizes rules and vocabulary of the languages to classify sentences. These techniques are highly accurate but susceptible to the problems of spelling errors and improper sentences. And these problems are frequent in any informal text including reviews and tweets. Also, in case of Hindi, [21] have trained a multinomial naive bayes model on annotated Hindi tweets to solve the problem.

Additionally, there have been efforts by researchers [19] to generate annotated resources by utilizing available raw corpus. They employ the availability of different domains to construct a Multi-arm Active Transfer Learning (MATL) algorithm to label raw samples and continuously add them to the original dataset. Each step updates the algorithm's parameters using reinforcement learning with a reward function. The above approach works well for the considered domains - sports, movies and politics. These domains have a formal vocabulary and grammar, whereas, tweets do not follow this trend. Hence, the model is inapplicable to unstructured tweets. The new resources depend on the available resources' domain, which is risky, especially in the case of tweets that do not comply with any certain domain.

Usually, methods that require the immutable words are ineffective. A better approach utilizes the languages' characters instead of words. Given their proven effectiveness in [8,6,24,1,9,13,25], we use Bidirectional LSTMs (Bi-LSTMs) based on character n-grams. This approach produces embeddings based on the sequence of character n-grams, thus eliminating the problems of spelling mistakes and agglutination (in the case of some languages such as Telugu).

Although Bi-LSTMs map the sentences to a sentiment space, we also require the distance between the sentences with the similar sentiment to be closer and the sentences with the different sentiment to be farther. For this reason, we use the architecture of Siamese Networks. This architecture possesses the capability of learning similarity from the given data without requiring specific information about the classes.

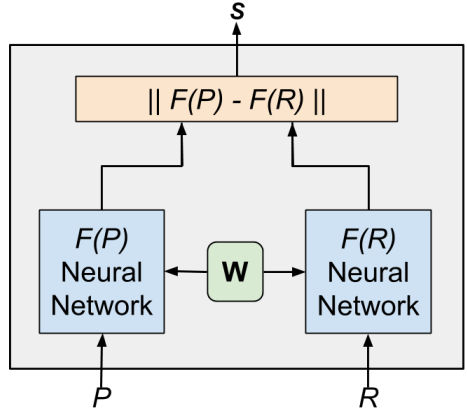

Figure 1: Siamese Network

## 2.1 Siamese Networks

[4] introduced siamese neural networks to solve the problem of signature verification. Later, [5] used the architecture with discriminative loss function for face verification. Recently, these networks solved the problem of community question answering [7]. Let, $F(X)$ be the family of functions with parameters $W$. $F(X)$ is differentiable with respect to $W$. Siamese network seeks a value of the parameter $W$ such that the symmetric similarity metric is small if $X1$ and $X2$ belong to the same category, and large if they belong to different categories. The scalar energy function $S(R, P)$ that measures the relatedness of sentiments between resource-poor ($P$) and resource-rich ($R$) language's tweets can be defined as:

$$S(P, R) = ||F(P) - F(R)|| \qquad (1)$$

In SNASA, we input the tweets from both the languages to the network. The loss function is minimized so that $S(P, R)$ is small if the $R$ and $P$ carry the same sentiment and large otherwise.

|  |  | 3 classes | 4 classes |  |  |  |
|---|---|---|---|---|---|---|
| Datasets | Sentence Length | Pos  Neg  Neu | V.Pos | Pos | Neg | V.Neg |
| English - Movie Reviews | 429 | 38%  24%  38% | 17% | 40% | 31% | 12% |
| English - Twitter | 12 | 29%  26%  45% | - | - | - | - |
| Spanish - Twitter | 14 | 48%  13%  39% | - | - | - | - |
| Hindi - Reviews | 15 | 33%  31%  36% | - | - | - | - |
| Telugu - News | 13 | 27%  27%  46% | - | - | - | - |

Table 1: Distribution of the datasets considered in the experiments. Pos, Neg, Neu, V. Pos and V.Neg stand for Positive, Negative, Neutral, Very Positive and Very Negative respectively. 4 classes are available only in Movie Review dataset.

## 3 Datasets

The datasets for different languages are given below:

- **English - Movie Review Dataset:** The dataset[20] consists of 5006 movie reviews annotated into 3 classes (positive, neutral and negative) and 4 classes (very positive, positive, negative and very negative).
- **English - Twitter Dataset:** The dataset[17] consists of 103035 tweets annotated into 3 classes - positive, neutral and negative.
- **Spanish - Twitter Dataset:** The dataset[17] consists of 275589 tweets annotated into 3 classes - positive, neutral and negative.
- **Hindi - Product Review Dataset:** The dataset[16] consists of 1004 product reviews annotated into 3 classes - positive, neutral and negative.
- **Telugu - News Dataset:** The dataset[19] is an annotated corpus of news data tagged into 3 classes - positive, neutral and negative.

The sentiment tags' distribution in the above datasets is given in Table 1.

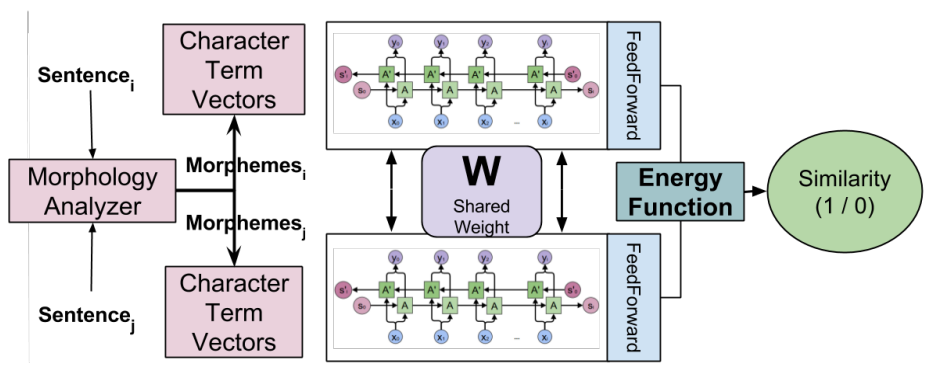

Figure 2: Architecture of SNASA

## 4 Architecture of SNASA

As shown in Figure 2, SNASA consists of a Bi-LSTMs pair and a dense feed forward layer at the top. The Bi-LSTMs capture the sequence and constituents of the sentence and project them to a sentiment space. We connect the yielded sentiment vectors to a layer that measures similarity between them. The contrastive loss function combines the similarity measure and the label. Back-propagation through time computes the loss function's gradient with respect to the weights and biases shared by the sub-networks.

| Language | Tel-News | Hin-Reviews | Spa-Twitter | Eng-Twitter | Eng-Movie Review |
|---|---|---|---|---|---|
| Char trigrams | 17424 | 6059 | 296797 | 197639 | 13897 |
| Words | 75417 | 10244 | 481280 | 359113 | 2148164 |

Table 2: Number of Unique Character Trigrams and Words in the datasets

### 4.1 Primary Representation

Informal data consists of a lot of spelling errors, out-of-vocabulary(OOV) words and multiple spelling of the same word. The way of writing a word may also convey a sentiment (e.g; "Hiiii" conveys a positive sentiment whereas "Hi" is a neutral sentiment). Hence, we use character trigrams to embed the sentence instead of using words. This approach takes care of the spelling errors and OOV words because a partial match exists in the character trigrams. Character trigrams take the information of all the inflections of a word, thus, eliminating the problem of agglutination. This method, also, captures the sentiment of different ways of writing as information is attained on a character-level. To further address the problem of agglutination in morphologically rich languages, we add a morphology analyzer that divides the words into its constituent morphemes. This also helps in the computational complexity as the number of character trigrams is far less than the number of complete words (shown in table 2). The approach represents a sentence using a vector with number of dimensions equal to the number of unique character trigrams in the training dataset.

We input character based term vectors of resource-poor and resource-rich language's tweets and a label to the twin networks of SNASA. The label indicates whether the samples are nearer or farther to each other in the sentiment space. For positive samples (nearer in the sentiment space), we feed the twin networks with term vectors of tweets (one from resource-poor and one from resource-rich) with the same sentiment tag. For negative samples (far away from each other in the sentiment space), we feed the twin networks with term vectors of tweets (one from resource-poor and one from resource-rich) with different sentiment tags.

### 4.2  Bi-directional LSTM Network

We map each sentence-pair into $[p_i, r_i]$ such that $p_i \in \mathbb{R}^m$ and $r_i \in \mathbb{R}^n$, where $m$ and $n$ are the total number of character trigrams in the resource-poor language and the resource-rich language respectively.

Bi-LSTM model encodes the sentence twice, one in the original order (forward) of the sentence and one in the reverse order (backward). Back-propagation through time [3] calculates the weights for both the orders independently. The algorithm works in the same way as general back-propagation, except in this case the back-propagation occurs over all the hidden states of the unfolded timesteps.

We, then, apply element-wise Rectified Linear Unit (ReLU) to the output encoding of the BiLSTM. ReLU is defined as: $f(x) = max(0, x)$. The choice of ReLU simplifies back-propagation, causes faster learning and avoids saturation.

The architecture's final dense feed forward layer converts the output of the ReLU layer into a fixed length vector $s \in \mathbb{R}^d$. In our architecture, we empirically set the value of $d$ to 128. The overall model is formalized as:

$$s = max\{0, W[fw, bw] + b\} \qquad (2)$$

where $W$ is a learned parameter matrix (weights), $fw$ is the forward LSTM encoding of the sentence, $bw$ is the backward LSTM encoding of the sentence, and $b$ is a bias term, then passed through an element-wise ReLU.

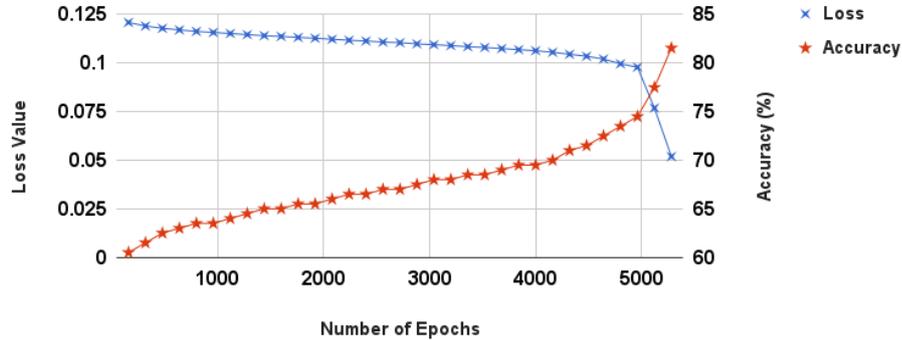

Figure 3: Number of Epochs vs Loss and Accuracy

## 5  Training and Testing

We train SNASA on the pairs of sentences in resource-poor and resource-rich language to capture their similarity in the sentiment. SNASA differs from other deep learning counterparts due to its property of parameter sharing. Training the network with a shared set of parameters not only reduces the number of parameters (thus, save many computations) but also ensures that the sentences of both the languages project into the same sentiment space. We learn the network's shared parameters to minimize the distance between the sentences with the same sentiment and maximize the distance between the tweets with different sentiment.

Given an input $p_i, r_i$ where $p_i$ and $r_i$ are tweets from resource-poor and resource-rich languages respectively and a label $y_i \in \{-1, 1\}$, the loss function is defined as:

$$l(p_i, r_i) = \begin{cases} 1 - cos(p_i, r_i), & y = 1; \\ max(0, cos(p_i, r_i) - m), & y = -1; \end{cases} \quad (3)$$

where $m$ is the margin that decides the distance by which dissimilar pairs should be moved away from each other. It generally varies between 0 to 1. The loss function is minimized such that pair of tweets with the label 1 (same emoji) are projected nearer to each other and pair of tweets with the label -1 (different emoji) are projected farther from each other in the sentiment space. The model is trained by minimizing the overall loss function in a batch. The objective is to minimize:

$$L(\Lambda) = \sum_{(p_i, r_i) \in C \cup C'} l(p_i, r_i) \quad (4)$$

where $C$ contains the batch of same sentiment sentence pairs and $C'$ contains the batch of different sentiment sentence pairs. Back-propagation through time (BPTT) updates the parameters shared by the Bi-LSTM sub-networks.

For testing, we randomly sample a certain number (100 in our case) of sentences for each sentiment $R_{sentiment}$ from the language corpus with higher amount of data. For every input, we then apply the trained model to get the similarity between the input and all corresponding $R_{sentiment}$. The $R_{sentiment}$ with the most matches with the input is finally selected as the correct polarity tag.

In case the correlated data available for both the resource-rich and resource-poor languages are not annotated, we use the one language's abundant resources to construct a state-of-the-art sentiment analysis model [10]. The sentiment analysis model in conjunction with the correlation data obtained from SNASA aids the resource-poor language's prediction.

## 6  Baselines

The approaches vary based on the language in consideration. Hence, baselines are also defined below accordingly. English, Japanese and Spanish enjoy the highest share of data on Twitter[1]. We consider English and Spanish because of their script and typological similarity (both are SVO). The baselines considered for resource-rich languages - English and Spanish are:

---

[1] The Many Tongues of Twitter - MIT Technology Review

|  | 3 classes | | | | 4 classes | | | |
|---|---|---|---|---|---|---|---|---|
| **Language Pair** | Accuracy | Precision | Recall | F-score | Accuracy | Precision | Recall | F-score |
| Eng-Eng | 81.25% | 0.83 | 0.80 | 0.81 | 66.1% | 0.67 | 0.64 | 0.65 |
| Eng-Hin | 80.5% | 0.82 | 0.79 | 0.80 | - | - | - | - |
| Eng-Tel | 80.3% | 0.82 | 0.79 | 0.80 | - | - | - | - |
| Eng-Spa | 81.5% | 0.83 | 0.80 | 0.81 | - | - | - | - |
| Hin-Tel | 70.2% | 0.72 | 0.69 | 0.70 | - | - | - | - |

Table 3: Comparison between different language pairs for 3 classes and 4 classes (only Movie Review).

- **Average Skip-gram Vectors (ASV):** We train a Word2Vec skip-gram model [15] on a corpus of 65 million raw sentences in English and 20 million raw sentences in Spanish. Word2Vec provides a vector for each word. We average the words' vectors to get the sentence's vector. So, each sentence vector is defined as:
$$V_s = \frac{\sum_{w \in W_s} V_w}{|W_s|} \quad (5)$$
where $V_s$ is the sentence's vector $s$, $W_s$ is the set of the words in the sentence and $V_w$ is the vector of the word $w$.
After obtaining each message's embedding, we train an L2-regularized logistic regression, (with $\epsilon$ equal to 0.001).
- **Matrix Vector Recursive Neural Network (MV-RNN):** The model[22] assigns a vector and a matrix to every node of a syntactic parsed tree. The vector represents the node's semantic value and the matrix represents its relation with the neighboring words. A recursive neural network model is then trained using backpropagation through structure to define the nodes' weighted contribution to the sentence's sentiment.
- **Adaboost based Convolutional Neural Network (Ada-CNN):** CNN sentence classifier models[12] with filter sizes 3,4 and 5 are trained on the datasets. These filter sizes capture the 3-gram, 4-gram and 5-gram contribution to the overall sentiment respectively. Adaboost then attains a weighted combination of these classifiers. This weighted combination of the classifiers assigns the overall sentiment tag. This helps in giving a weighted emphasis to the information provided by 3-grams, 4-grams and 5-grams in the sentence.

Hindi and Telugu are the $3^{rd}$ and $17^{th}$ most spoken language in the world respectively. But they hold a relatively low share of Twitter data. The speakers of Hindi and Telugu on Twitter primarily use the roman transliterated form of the language. This also further translates to a limited availability of annotated corpus for these languages. The baselines for these languages are:

- **Domain Specific Classifier (Telugu) (DSC-T):** We train a Word2Vec model on a corpus of 700,000 raw Telugu sentences provided by Indian Languages Corpora Initiative (ILCI). We train a Random Forest (RF) and Support Vector Machines classifier (SVM) (given by [18]) on the Telugu News dataset to construct our baseline for Telugu language.

|  | 3 classes | | | | 4 classes | | | |
| --- | --- | --- | --- | --- | --- | --- | --- | --- |
| **Method** | Accuracy | Precision | Recall | F-score | Accuracy | Precision | Recall | F-score |
| ASV | 52.59% | 0.49 | 0.52 | 0.50 | 39.42% | 0.47 | 0.45 | 0.32 |
| MV-RNN | 79.0% | 0.77 | 0.75 | 0.76 | 64.3% | 0.63 | 0.62 | 0.62 |
| SNASA | **81.25%** | **0.83** | **0.80** | **0.81** | **66.1%** | **0.67** | **0.64** | **0.65** |

Table 4: Comparison with the English baselines on Movie Review dataset. ASV is Average Skip-gram Vectors, MV-RNN refer to Matrix Vector Recursive Neural Network model.

| **Method** | Accuracy | Precision | Recall | F-score |
| --- | --- | --- | --- | --- |
| ASV | 52.59% | 0.49 | 0.52 | 0.50 |
| MV-RNN | 79.0% | 0.77 | 0.75 | 0.76 |
| DSC-T | 68.17% | 0.67 | 0.66 | 0.66 |
| MNB-H | 62.14% | 0.61 | 0.58 | 0.59 |
| SNASA (Eng-Eng) | **81.25%** | **0.83** | **0.80** | **0.81** |
| SNASA (Hin-Eng) | **80.5%** | **0.82** | **0.79** | **0.80** |
| SNASA (Tel-Eng) | **80.3%** | **0.82** | **0.79** | **0.80** |

Table 5: Comparison with the baselines on three-class datasets of the respective languages. ASV is Average Skip-gram Vectors, MV-RNN refer to Matrix Vector Recursive Neural Network model. They are baselines for English and compare to SNASA (Eng-Eng). DSC-T is Domain Specific Classifier for Telugu and compares to SNASA (Tel-Eng). MNB-H refers to Multinomial Bayes Model for Hindi and compares to SNASA (Hin-Eng).

– **Multinomial Naive Bayes Model (Hindi) (MNB-H):** We train a multinomial naive bayes model (given by [21]) on the Hindi Review dataset to form our baseline for Hindi language.

## 7 Experiments and Evaluation

In order to study the comparison of SNASA to the previous models, we performed an array of experiments. In the first experiment (section 7.1), we analyze varying language pairs and make a comparison between them. In the second experiment (section 7.2), we compare our model against previous approaches in the problem of Sentiment Analysis. In the third experiment (section 7.3), we provide an extension where emojis retrieved from Twitter are utilized instead of regular sentiment tags.

### 7.1 Experiments for different language pairs

The experiment is a classification task. We take the English and Hindi three-class datasets (Eng-Hin) and align each Hindi sentence with English sentences of the same sentiment (positive samples) and label them 1. Similarly, we also randomly sample the same number of English tweets with different sentiment (negative samples) for each Hindi Tweet and label them -1.

Similarly, we repeat the experiment for English-Telugu (Eng-Tel) dataset pair, English-Spanish (Eng-Spa) dataset pair, English-English (Eng-Eng) dataset pair and Hindi-Telugu (Hin-Tel) dataset pair. Table 3 demonstrates the results of the experiments.

We run another experiment for the case of English (Eng-Eng), where we take the case of Movie Review dataset and align each sentence with other sentences of the same sentiment (positive samples) and label them 1. Similarly, we also randomly sample the same number of sentences with different sentiment (negative samples) and label them -1. We perform the experiment for both three-class and four-class classification task. The results of this experiment are given in Table 3.

### 7.2 Comparison with the Baselines

In this experiment, we compare our model against the baselines (defined in section 6).

We defined the baselines for resource-rich languages on English. So, we perform contrastive learning of our model using data made by aligning each English sentence with a set of positive samples (with the same sentiment) with label 1 and a set of negative samples (with different sentiment) of the same size with label -1.

In the case of resource-poor languages, i.e. Hindi and Telugu, we perform contrastive learning of our model using data made by aligning each of the resource-poor language (Hindi and Telugu) sentence with a set of positive English samples (with the same sentiment) with label 1 and a set of negative English samples (with different sentiment) of the same size with label -1.

The baselines on English are trained and evaluated on both Movie Review dataset and three-class dataset. The baselines on Spanish, Hindi and Telugu are trained and evaluated on their respective three-class datasets.

The results of the comparison between SNASA and previous approaches on Movie Review dataset are given in Table 4. The comparison between SNASA and previous approaches on three-class datasets are given in Table 5.

| Emojis | Class | Eng | Spa | Hin | Tel |
|---|---|---|---|---|---|
| ❤️😍😁😊 | Positive | 37% | 36% | 39% | 39% |
| 😐🤔😶😏 | Neutral | 31% | 30% | 31% | 31% |
| 😠😟😖😡 | Negative | 32% | 34% | 30% | 30% |

Table 6: Distribution after mapping Emojis to respective sentiment classes.

|  | SNASA | | Emoji-SNASA | |
|---|---|---|---|---|
| Dataset | A(%) | F1 | A(%) | F1 |
| English | 81.25% | 0.81 | **84.8%** | **0.83** |
| Spanish | 81.5% | 0.81 | **85.2%** | **0.83** |

Table 7: Performance enhancement due to emojis in sentiment analysis.

### 7.3 Emoji based approach with SNASA (Emoji-SNASA)

In our previous experiment (section 7.1), we found that in several test scenarios, the tweet is incorrectly classified because of limited correlation data available between the language pair. Emojis are characters used in social media to communicate context inexpressible by normal characters. A major application of these emojis is in expressing sentiment. So, we use the emojis available in our datasets to align language pairs instead of sentiment tags. The emojis in the dataset are classified manually into sentiment classes by three annotators. The emojis were taken into consideration only if all the three annotators were in agreement. The distribution of each of thus formed sentiment classes is given in Table 6.

We align each English sentence with a set of positive samples (with the same emoji) with label 1 and a set of negative samples (with different emoji) of the same size with label -1. The results for the experiment are given in Table 7.

### 7.4 Evaluation of the Experiments

We observe from Table 3 that the best overall results for sentiment analysis are seen for the English-Spanish pair. This is due to the English-Spanish containing the maximum number of tweet pairs. We also note from Figure 3 that with increasing number of epochs, the accuracy and overall performance considerably gets better.

Multiple times a sentence is misclassified because of incorrect correlation between the languages in the pair. We corrected this behavior using emojis in three-class datasets to increase the number of sentences that could be used to establish correlation. To verify this behavior, we conducted another experiment in section 7.3 to approach this from the perspective of emojis instead of sentiment tags. The experiment's result (given in Table 7) demonstrate that emojis lead to better accuracy. This is seen because emojis lead to a better correlation between the languages' pair. However, emojis do not always represent perfect sentiment and hence will increase the performance only if the data taken has limited noise.

From table 4 and 5, we observe that SNASA outperforms the current approaches significantly, especially in the case of resource-poor languages. Interestingly, the results also show that using shared parameters leads to an improvement in performance. SNASA learns representation, specifically, for the task of sentiment classification. It leverages the relatively resource-rich language for the improvement in the resource-scarce language's performance.

## 8   Conclusions

In this paper, we proposed SNASA for sentiment analysis of resource-poor languages which solves the problem by projecting the resource-poor language and resource-rich language in the same sentiment space. SNASA employs twin Bidirectional LSTM networks with shared parameters to capture a sentiment based representation of the sentences. These sentiment based representations are used in conjunction with a similarity metric to group sentences with similar sentiment together.

An emoji based approach used in conjunction with SNASA boosts the performance of overall sentiment analysis further. Experiments conducted on three-class and four-class (Movie Review) datasets revealed that SNASA outperforms the current state-of-the-art approaches significantly.

In future, we would like to apply the current model on more applications based on learning similarity like question-answering, conversation systems and semantic similarity. Though, of course, the presence and impact of correlation between languages would be limited in other areas. Also, we believe that there is a good case for integration of attention-based models in the subnetworks.